\documentclass[letterpaper, 10 pt, conference]{source/ieeeconf}  

\IEEEoverridecommandlockouts                              

\usepackage{graphics} 
\usepackage{epsfig} 
\usepackage{times} 
\usepackage{amsfonts,amssymb,amsmath,bm}
\usepackage{mathtools}
\usepackage{amstext}
\usepackage{ifthen}
\usepackage[english]{babel}
\usepackage{algorithm}
\usepackage{algorithmic}
\usepackage{siunitx,booktabs}
\usepackage[acronym]{glossaries}
\usepackage{cite}
\usepackage[hidelinks]{hyperref}
\usepackage{comment}
\usepackage{subcaption}
\usepackage[capitalise]{cleveref}
\usepackage[font=small,labelfont=bf,tableposition=top]{caption}
\usepackage{stfloats}

\let\ACMmaketitle=\maketitle
\renewcommand{\maketitle}{\begingroup\let\footnote=\thanks \ACMmaketitle\endgroup}

\makeatletter
\newcommand*\titleheader[1]{\begingroup\gdef\@titleheader{#1}\let\footnote=\thanks\endgroup}
\AtBeginDocument{%
  \let\st@red@title\@title
  \def\@title{%
  \begin{flushleft}
    \vspace{-2.0em}
    \bgroup\normalfont\small\@titleheader\par\egroup
    \vspace{-18pt}\par\noindent\rule{\textwidth}{0.1pt}
    \end{flushleft}
    \vskip0.5em\st@red@title
        }

}

\title{\LARGE \bf
Hierarchical Policy Blending as Inference for Reactive Robot Control}

\author{Kay Hansel$^{1}$,  Julen Urain$^{1}$, Jan Peters$^{1-4}$ and Georgia Chalvatzaki$^{1,3}$
\thanks{This work received funding by the DFG project CHIRON and the Emmy Noether Programme (CH 2676/1-1), and the EU project ShareWork.}
\thanks{$^{1}$ Computer Science Department, Technische Universität Darmstadt (Germany), $^2$ German Research Center for AI (DFKI), $^3$ Hessian.AI, $^4$ Centre for Cognitive Science {\tt\footnotesize\{kay.hansel, julen.urain, jan.peters, georgia.chalvatzaki\} @tu-darmstadt.de}}%
\thanks{This work has been submitted to the IEEE for possible publication. Copyright may be transferred without notice, after which this version may no longer be accessible.}%
}
\titleheader{\textit{Preprint}}
\newcommand{\mc}[1]{\mathcal{#1}}

\DeclareMathOperator*{\argmax}{\arg\!\max}
\DeclareMathOperator*{\argmin}{\arg\!\min}

\DeclarePairedDelimiterX{\doubleVertBar}[2]{[}{]}{%
  #1\;\delimsize\|\;#2%
}
\newcommand{\KL}{\mathbb{D_{\text{KL}}}\doubleVertBar}

\renewcommand{\H}[2][]{
\ifthenelse {\equal{#1}{}}
{\mathbb{H}\left[#2\right]}
{\mathbb{H}_{#1}\left[#2\right]}}

\newcommand{\E}[2][]{
\ifthenelse {\equal{#1}{}}
{\mathbb{E}\left[#2\right]}
{\mathbb{E}_{#1}\left[#2\right]}}

\newcommand{\Sspace}{\mc{S}}
\newcommand{\Aspace}{\mc{A}}

\newcommand{\nexperts}{n}
\newcommand{\nobstacles}{m}

\newcommand{\state}{\vs}
\newcommand{\action}{\va}

\newcommand{\parameter}{\vtheta}
\newcommand{\tstep}{t}
\newcommand{\horizon}{h}

\def\1{\bm{1}}

\def\RR{\mathbb{R}}

\def\rd{{\textnormal{d}}}

\def\vmu{{\bm{\mu}}}
\def\vtheta{{\bm{\theta}}}
\def\va{{\bm{a}}}

\def\vc{{\bm{c}}}

\def\vq{{\bm{q}}}

\def\vs{{\bm{s}}}

\def\vbeta{{\boldsymbol{\beta}}}

\def\vmu{{\boldsymbol{\mu}}}

\def\vtheta{{\boldsymbol{\theta}}}
\def\vtau{{\boldsymbol{\tau}}}

\def\mJ{{\bm{J}}}

\def\mLambda{{\bm{\Lambda}}}

\DeclareMathAlphabet{\mathsfit}{\encodingdefault}{\sfdefault}{m}{sl}
\SetMathAlphabet{\mathsfit}{bold}{\encodingdefault}{\sfdefault}{bx}{n}

\def\gJ{{\mathcal{J}}}

\def\gN{{\mathcal{N}}}
\def\gO{{\mathcal{O}}}

\def\gQ{{\mathcal{Q}}}

\def\gX{{\mathcal{X}}}

\makeatletter
\newcommand{\StatexIndent}[1][3]{%
  \setlength\@tempdima{\algorithmicindent}%
  \Statex\hskip\dimexpr#1\@tempdima\relax}
\makeatother
\newacronym{poe}{PoE}{product of experts}
\newacronym{kld}{KL divergence}{Kullback–Leibler divergence}

\newacronym{map}{MAP}{maximum a posteriori}
\newacronym{rmp}{RMP}{Riemannian motion policies}
\newacronym{rmpflow}{RMP\textit{flow}}{Riemannian motion policies}
\newacronym{mpcicem}{iCEM-MPC}{improved cross entropy method for model-predictive-control}
\newacronym{icem}{iCEM}{improved cross entropy method}
\newacronym{mpc}{MPC}{model predictive control}
\newacronym{hipbi}{HiPBI}{hierarchical policy blending as inference}

\newacronym{2dmaze}{2D-Maze}{2D toy maze environment}
\newacronym{2dbox}{2D-Box}{2D toy box environment}
\begin{document}
\maketitle
\thispagestyle{empty}
\pagestyle{empty}
\begin{abstract}
Motion generation in cluttered, dense, and dynamic environments is a central topic in robotics, rendered as a multi-objective decision-making problem. 
Current approaches trade-off between safety and performance.
On the one hand, reactive policies guarantee a fast response to environmental changes at the risk of suboptimal behavior. 
On the other hand, planning-based motion generation provides feasible trajectories, but the high computational cost may limit the control frequency and, thus, safety.
To combine the benefits of reactive policies and planning, we propose a hierarchical motion generation method.
Moreover, we employ probabilistic inference methods to formalize the hierarchical model and stochastic optimization. 
We realize this approach as a weighted product of stochastic, reactive expert policies, where planning is used to adaptively compute the optimal weights over the task horizon.
This stochastic optimization avoids local optima and proposes feasible \emph{reactive} plans that find paths in cluttered and dense environments. 
Our extensive experimental study in planar navigation and 7DoF manipulation shows that our proposed hierarchical motion generation method outperforms both myopic reactive controllers and online re-planning methods. 
Additional material available at \url{https://sites.google.com/view/hipbi}.
\end{abstract}
\section{Introduction}
We expect autonomous general-purpose robots to navigate in unstructured, cluttered environments and perform multiple tasks autonomously.
The robots need to guarantee the success of the task while responding reactively and safely to dynamic changes in the environment~\cite{khatib1987unified,park2008movement,hogan2012dynamic,ijspeert2013dynamical,paraschos_2018_promp,ratliff_rmp_2018}.
Robot motion generation encompasses all methods that define control commands to generate coordinated robot behaviors.
These control commands take either the form of trajectory-level control~\cite{zucker_2013_chomp,kalakrishnan_2011_stomp,lambert2022learning} or direct, instantaneous controls such as joint velocity and acceleration~\cite{khansari-zadeh_2011_seds,cheng_rmpflow_2018, xie_gf_2020, ratliff_gf_2021, van_gf_2022}.

\begin{figure}[ht!]
    \centering
    \includegraphics[width=1.\linewidth]{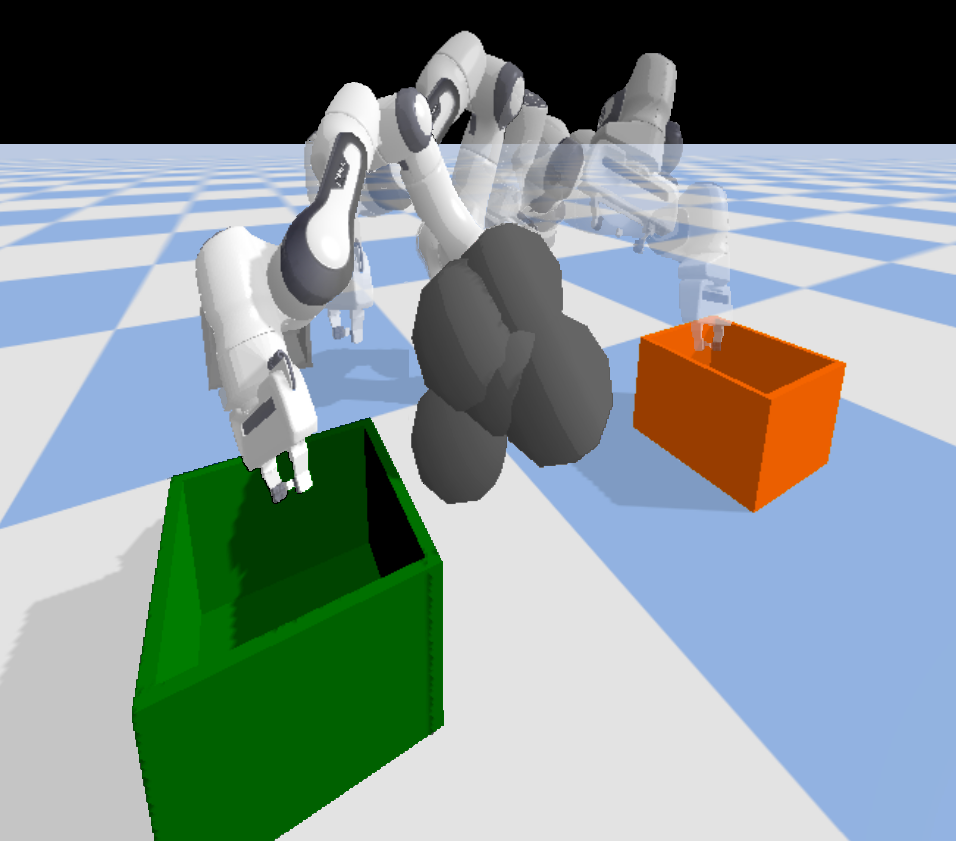}
    \caption{A sequence of the reactive motion of a 7DoF manipulator robot. The robot starts moving from the orange box toward the green box. Our proposed method enables a reactive motion that avoids collisions with the grey obstacle and overcomes local minima resulting from multiple constraints.}
    \vspace{-0.6cm}
    \label{fig:main_fig}
\end{figure}

On one side of the spectrum are motion planning approaches. 
Motion planning can be categorized into two subgroups based on the underlying methodology. 
First, sample-based methods provide probabilistic completeness guarantees in terms of goal-reaching and collision-free paths~\cite{kavraki_1996_probabilistic, lavalle_1998_rapidly, lavalle_2006_planning,kalakrishnan_2011_stomp, low_2021_prompt}. 
Gradient-based trajectory optimization methods~\cite{zucker_2013_chomp, mukadam_2018_gpmp}, however, search for the optimal path to a task-specific goal, taking into account several underlying objectives, e.g., trajectory smoothness, obstacle avoidance, or joint limits avoidance. Algorithms from these two categories consider a static environment with a predefined and static goal location. 
These approaches are, thus, specifically suited for robot operations in structured environments.
To enhance adaptability in dynamic environments, online planners~\cite{kobilarov_cemmpc_2012, pinneri_icem_2021,bhardwaj_storm_2022,williams_mppi_2017,lambert_svmpc_2020}, attempt to ensure reactivity through online re-planning over the action space for a short planning horizon. 
These methods find the optimal sequence of actions for a certain planning horizon. 
The first proposed action of the sequence is executed, followed by re-planning. 
While some efforts have been made to adapt these approaches to high-dimensional robots~\cite{bhardwaj_storm_2022}, these methods become computationally intensive in high dimensions, limiting the maximum frequency of control and, thus, reactivity.

Reactive motion generators are on the other side of the spectrum. 
These methods aim to meet high-frequency requirements to provide instantaneous control. 
Therefore, reactive motion generators ensure adaptability to environmental changes~\cite{khatib1986real,calinon2018robot} and hence local safety, e.g., by repulsive potential fields for obstacle avoidance. 
However, these methods do not provide look-ahead planning due to the high-frequency requirements.
As a result, reactive motion generators are myopic.
The myopic behavior makes the reactive methods susceptible to getting trapped in local minima when solving simultaneous objectives, such as obstacle and self-collision avoidance, as well as goal-reaching.



In this work, we propose a \gls{hipbi} approach that employs hierarchical decision making and probabilistic inference to combine the benefits of online motion planning and reactive motion generators.
On the lower level, we apply \gls{rmp}~\cite{ratliff_rmp_2018, cheng_rmpflow_2018}. \gls{rmp}s represent myopic policies and provide a high-frequency response to environmental changes. 
On the higher level, we adopt planning-as-inference and use a sampling-based look-ahead planner that operates on the parameter space of the underlying \gls{rmp}s.
Therefore, we reformulate the \gls{rmp}s as Gaussian policies to employ planning-as-inference methods~\cite{toussaint_2009_inference, levine_maxentrl_2018} resulting in a \gls{poe}~\cite{tresp_2000_bcm, hinton_cd_2002} formulation that belongs to the exponential family.
To evaluate the performance of our approach, we conducted empirical studies for 2DoF point-mass navigation and 7DoF robot manipulation tasks in complex and dynamic environments.
We compared our approach with representative reactive motion generation baselines. 
Our results highlight the effectiveness of \gls{hipbi} in solving tasks faced with cluttered and dynamic obstacles.  
By combining high-level planning with low-level reactive control, our approach achieves high success and safety rates.

\paragraph*{\textbf{Contribution}}
Our contributions are: 
(i) Developing a robot motion generation method that hierarchically combines reactive motion generation and sample-based online planning; 
(ii) Employing planning-as-inference to address the policy blending problem in an online fashion; 
(iii) Empirically demonstrating the achievement of feasible and reactive motions.

\section{RELATED WORK}
%
Hierarchical decision-making -- abstracting the motion generation problem into multiple decision levels -- is well-known practice in robotics.
Such methods rely on multi-level planners or operate in the parameter space of motion policies. 
The former, such as TAMP, hierarchical planning, or hierarchical RL~\cite{kaelbling2010hierarchical,srivastava2014combined, jauhri_2022_mm, pertsch2020long}, generate subgoals that an underlying planner or policy must achieve.
The latter either specifically adjusts constraint functions of dynamic motion primitives~\cite{bahl2020neural,bahl2021hierarchical} or selects a policy from a mixture of experts~\cite{daniel2012hierarchical,kroemer2015towards,end2017layered,celik2022specializing,akrour2021continuous,zaki_2022_improperRL}.
Given the multi-centric nature of robotic tasks, a hierarchical mixture of experts selects only one of the experts~\cite{celik2022specializing}.
When faced with unexpected environmental changes, this selective behavior leads to suboptimal performance~\cite{kroemer2015towards} or at least to a combination of experts with already complexly encoded behaviors, e.g., by imitation or reinforcement learning~\cite{celik2022specializing}.
We argue that the composition of simple and stable reactive policies is capable of generating complex reactive behaviors in robotics~\cite{urain_cep_2021, li_rmp2_2021, mukadam_2020_rmp}.

The fundamental work for reactive motion generators applied artificial potential fields for modeling obstacle avoidance (repulsive) and goal-reaching (attractive) behavior~\cite{khatib1987unified}. 
This work formed the basis for operational space control investigating reactive policies to achieve instantaneous robot motion control~\cite{nakanishi2008operational, toussaint_2010_bayesian}. 
Using Riemannian metrics instead of Euclidean, the \gls{rmp} framework extends operational space control considering geodesics near obstacles~\cite{ratliff_rmp_2018, cheng_rmpflow_2018, mukadam_2020_rmp, li_rmp2_2021}. 
The recently proposed geometric fabrics generalize RMPs by employing Finsler geometry rather than Riemannian one~\cite{xie_gf_2020, ratliff_gf_2021, van_gf_2022}. 
Motion primitives provide learning-based generation of reactive, stable behavior~\cite{khansari-zadeh_2011_seds, ijspeert2013dynamical, calinon2014robot}. 
While most motion generators promise to be locally reactive, they are prone to get trapped in local minima.

\gls{rmp}'s intuitive superposition of policies corresponds to a product of experts~\cite{cheng_rmpflow_2018, mukadam_2020_rmp}. 
In the context of primitives, blending is useful to express complex behavior out of previously encoded primitives. Therefore, several works applied blending for parameterized motion primitives~\cite{luksch_2012_blendingprimitives, saverino_2019_mergingprimitives, paraschos_2018_promp} or Gaussian processes~\cite{cao_2014_gpoe, cao_2015_gpoe, deisenroth_2015_dgp, liu_2018_rbcm, cohen_2020_gpe}. 
Recent work addressed the blending problem utilizing QP optimization~\cite{jaquier_2022_blending}, energy-based models~\cite{urain_cep_2021, lambert2022learning}, or learning from demonstrations~\cite{pignat_2022_poe}. 
However, most methods assume equal importance for all experts or adjust the importance offline through optimization or learning.

\section{PRELIMINARIES}
This section introduces the necessary background to frame the hierarchical model and stochastic optimization. First, we present a concept to express optimal control as a Bayesian inference problem. Second, we provide a mathematical object to describe the expert policies for reactive motion generation.
\paragraph*{\textbf{Control as Inference}}
Denoting the system state $\vs_t \in \RR^s$ and action $\va_t\in \RR^a$ at time instant $t$, we define a discrete-time state-action trajectory as the sequence $\vtau \triangleq (\vs_0, \va_0, \vs_1, \va_1, \dots, \vs_T, \va_T)$ over a time horizon $T$. Given the transition dynamics ${p(\vs_{t+1}\mid\vs_t, \va_t)}$ and a policy ${\pi(\va_t\mid \vs_t; \vtheta)}$ conditioned on the parameters $\vtheta$, in optimal control, we aim to find the policy parameters $\vtheta$ that minimize the expected objective function $\gJ_\text{C}(\vtau)$,
where
${\vtheta^* = \arg \min_{\vtheta} \E[p', \pi]{ \gJ_\text{C}(\vtau;\vtheta)}.}$

We express the objective function ${\gJ_\text{C}(\vtau) = \sum_k c_k(\vtau)}$ as the sum of task-related cost functions, e.g., trajectory smoothness, collision avoidance, distance to target. 
Optimal control can be framed as a Bayesian inference problem considering the distribution over the policy parameters $\vtheta$~\cite{botvinick2012planning, rawlik2012stochastic, levine_maxentrl_2018}. We introduce a binary random variable $\gO \in \{0,1\}$ that indicates the optimality of a trajectory $\vtau$ w.r.t. the objective function $\gJ_\text{C}(\vtau)$. We express the probability of $\gO{=}1$ given a trajectory $\vtau$ by ${p(\gO{=}1|\vtau) \propto \exp(-\gJ_\text{C}(\vtau))}$. Given the trajectory distribution
\begin{align*}
    p(\vtau |\vtheta) = p(\vs_0)\,
    \textstyle\prod_{t=0}^{T-1}p(\vs_{t+1}|\vs_{t},\va_{t})\,\pi(\va_t|\vs_t;\vtheta),
\end{align*}
we can infer the optimality likelihood $p(\gO{=}1|\vtheta)$ as the marginal probability over all the state and action trajectories,
\begin{align*}
    p(\gO{=}1|\vtheta) = \textstyle\int_{\vtau} p(\gO=1|\vtau)
    \,p(\vtau|\vtheta)\,\rd\vtau.
\end{align*}
Note that ${\log p(\gO{=}1|\vtheta) \propto -\gJ_\text{C}(\vtau;\vtheta)}$.
Given a prior distribution $p(\vtheta)$, we can approximate the posterior distribution given the optimality $\gO=1$ as
${q(\vtheta|\gO{=}1) \propto p(\gO{=}1|\vtheta)\, p(\vtheta)},$
that updates the distribution over $\vtheta$ towards the parameters that are optimal for our objective function $\gJ_\text{C}(\vtau)$.
\paragraph*{\textbf{Riemannian Motion Policies}}
\gls{rmp}s~\cite{ratliff_rmp_2018, cheng_rmpflow_2018,li_rmp2_2021} describe a mathematical object for representing reactive, modular motion generation policies. In \gls{rmp}s, the action $\va_t$ at time instant $t$ is computed by a weighted sum of a set of policy components, $\va_t = \sum_i \alpha_i(\vs_t) \pi_i(\vs_t)$, 
with $\alpha_i>0$ the weighting term for the component $i$.
In \gls{rmp}s, the state ${\vs_t = (\vq_t, \dot{\vq}_t, \vc_t)}$ represents the robot's position $\vq_t\in \RR^q$, velocity $\dot{\vq}_t\in \RR^q$ and environment context $\vc_t$.
The action is chosen to be the robot acceleration $\va_t=\ddot{\vq}_t \in \RR^q$. We assume a set of task maps $\phi: \gQ \xrightarrow{} \gX$, that relate the robot configuration $\gQ$ space and a certain task space $\gX$. Then, given a task-space policy $\pi_x$, we can represent a policy in the robot configuration space by $\pi_q = \mJ_{\phi}^{\dagger} \pi_x(\phi(\vs_t))$, with $\mJ_{\phi}^{\dagger}$ the Jacobian pseudoinverse of the task map $\phi$.




\section{Hierarchical Reactive Policy Blending}
We propose a hierarchical optimization framework to blend reactive policies for motion generation in complex and dynamic environments. 
Our approach, \gls{hipbi}, expresses the blended policy as a \gls{poe}. A weighted superposition of the hand-tuned reactive policies determines the optimal action at the low level. At the high level, we formalize the optimization of the weights as a probabilistic inference problem and employ a sampling-based look-ahead planner. The hierarchical optimization scheme offers (i) high-frequency control and, hence, fast adaptation to environmental changes; (ii) avoidance of local optima in cluttered and dense environments through the deliberate superposition of policies. In the following, we will discuss both sublevels sequentially.

\paragraph*{\textbf{Weighted superposition of reactive policies}}
We adopt a probabilistic viewpoint and formalize each policy component of the \gls{rmp} as an energy-based model
\begin{align*}
	\pi_{i}(\va_t \mid \vs_t; \vtheta_{i})
    \propto
    \exp(-E_i(\vs_t, \va_t; \vtheta_{i})),
\end{align*}
taking the form of a Boltzmann distribution. The quantities $\vs_t \in \Sspace$ and $\va_t \in \Aspace$ denote a state and action at time instance $t$, respectively. An energy function ${E_i:\Sspace \times \Aspace \rightarrow \RR}$ assigns a cost to each state-action pair. Thus, the Boltzmann distribution gives a probability value to each state-action pair. The choice of the energy function $E_i$ and its hyperparameter $\vtheta_{i}$ is usually made in advance. In the case of Riemannian geometry, the energy $E_{i}(\vs_t, \va_t; \vtheta_{i})$ is a quadratic function satisfying smoothness and convexity. Accordingly, the Boltzmann distribution forms a multivariate Gaussian 
${\pi_{i}(\va_t \mid \vs_t; \vtheta_{i}) = \gN(\vmu_i(\vs_t), \mLambda_i(\vs_t)^{-1})}$
with $\vmu_i(\vs_t)$ and $\mLambda_i(\vs_t)$ as the mean and the precision matrix, respectively~\cite{urain_cep_2021}. 
Referring to \gls{rmp}s, the mean is the forcing function, and the precision matrix's inverse corresponds to the Riemannian metric.
We leverage a \gls{poe}
\begin{align}
\label{eq:3:poe}
    \pi(\va_t\mid \vs_t, \vbeta) \propto \textstyle\prod_{i=1}^{\nexperts}
    \,
    \pi_{i}(\va_t \mid \vs_t; \vtheta_{i})^{\beta_i},
\end{align}
with weighting factors $\vbeta$, also known as \textit{temperatures}, representing the importance or relevance of each policy in the product. In the logarithmic space, this blending equals a weighted superposition. 
The optimal action at time instance $t$ results from $\va_t^* = \argmax_{\va\in \Aspace} \log \pi (\va \mid \vs_t, \vbeta),$ depending on state $\vs_t$ and the weights $\vbeta$.
Due to the quadratic nature of the defined energy functions $E_i(\vs_t, \va_t)$ from the \gls{rmp} framework, the \gls{poe} is a Multivariate Gaussian distribution.
Therefore, we can obtain the gradient and hence the optimal action analytically in closed form. 

\paragraph*{\textbf{Sampling-based online planner}}
The behavior of the agent ${\pi(\vs_t, \va_t)}$ and hence an applied action sequence ${\va_{t}^*,\dots, \va_{t+h}^*}$ up to time $t+h$ are induced by the superposition of $\nexperts$ experts as in \eqref{eq:3:poe}. The temperature values $\vbeta$ give us the possibility to change the relevance or importance of an expert. In an online fashion, a change in the relevance of experts makes it possible to induce planning into the myopic nature of the policy ${\pi(\vs_t, \va_t)}$.
With the formulation of \gls{poe} in mind, we exploit the duality between control and probabilistic inference by formalizing an optimization procedure for deriving optimal weights. The posterior of optimal blending is given 
\begin{align*}
    p(\vbeta \mid \gO{=}1, \state_t) \propto
    p(\gO{=}1 \mid \vbeta , \state_t)\,
    p(\vbeta \mid \state_t),
\end{align*}
with the current state $\vs_t$ and the optimal likelihood ${p(\gO{=}1\mid \vbeta, \vs_t)}$ \cite{toussaint_2009_inference, levine_maxentrl_2018}. Given this likelihood, we can insert desired higher-level goals into the framework to achieve a desired behavior. 
Assuming a parameterized variational distribution ${q(\vbeta; \vtheta)}$, we minimize the reverse \acrlong{kld}
\begin{align*}
    q^{*}(\vbeta; \vtheta) = \textstyle\argmin_{q(\vbeta; \vtheta)} \KL{q(\vbeta; \vtheta)}{p(\vbeta \mid \gO{=}1, \state_{t})}.
\end{align*}
The reverse \acrshort{kld} ensures that the optimization fits the distribution ${q(\vbeta; \vtheta)}$ to modes of the ${p(\vbeta \mid \gO{=}1, \state_{t})}$. Due to the fact that there are potentially many optimal solutions, the mode-seeking behavior is beneficial. The optimal distribution ${q^{*}(\vbeta; \parameter)}$ is therefore obtained from
\begin{align*}
   q^{*}(\vbeta; \parameter) = \min_{q(\vbeta; \parameter)}\left[\H[q(\vbeta, \parameter)]{\textstyle\frac{p(\gO{=}1 \mid \vbeta, \state_t)\,p(\vbeta \mid \state_t)}{q(\vbeta; \parameter)}}\right].  
\end{align*}
Given the inference framework, we can impose important properties on the temperature parameters. Using an exponential distribution as ${q(\vbeta; \parameter)}$, we can ensure that all weights are greater than zero. Additionally, the use of a Dirichlet distribution implies that the temperatures sum to one. We choose $q(\vbeta; \parameter)=\mathrm{Dir(\vbeta; \parameter)}$ for this reason. The prior ${p(\vbeta \mid \vs_t)}$ gives us the opportunity to incorporate prior knowledge into the framework. As we do not want to bias the optimization, we consider a uniform prior. The Shannon entropy of the variational distribution ${q(\vbeta; \vtheta)}$ itself ensures that the distribution does not collapse.
The  optimal likelihood forms a marginal likelihood 
\begin{align*}
    p(\gO{=}1 \mid \vbeta , \vs_{t}) = \textstyle\int p(\gO{=}1 \mid \vtau, \state_{1})\,p(\tau\mid \vbeta , \vs_{1}) \,\rd\tau,
\end{align*}
over possible trajectories $\tau$. Inferring this quantity is challenging due to the integral over $\vtau$ on the right hand side. 
We employ variational inference that defines a lower bound, also known as evidence lower bound (ELBO) \cite{beal_2003_variational, tzikas_2008_variational}. Hence, we choose a variational distribution 
\begin{align*}
    \hat{q}(\vtau\mid\vbeta) = \hat{q}(\state_t)
    \,\textstyle\prod_{i=\tstep}^{\horizon}\hat{q}(\state_{t+1} \mid \va_t, \vs_{t})\,\pi(\va_t\mid \vbeta, \vs_{t}),
\end{align*}
over $\vtau$. Assuming that ${\hat{q}(\vtau\mid\vbeta)}$ approximates the true trajectory distribution ${p(\vtau \mid \vbeta, \vs_{t})}$ closely - given $\vbeta$ and $\vs_t$ - we obtain the ELBO
\begin{align}
    \label{eq:3:lower_bound}
    \E[q(\vbeta)]{\E[\hat{q}(\tau)]{\textstyle\sum_{i=t}^{h}\log\frac{ p(\gO_{t}{=}1\mid, \action_{t}, \state_{t})}{\pi(\va_t\mid \vbeta, \vs_{t})}}
    + \log q(\vbeta)}.
\end{align}
For clarity, we omitted the dependencies on $\vtheta$ and $\vbeta$ of ${q(\vbeta\mid\vtheta)}$ and ${\hat{q}(\tau\mid\vbeta)}$. 
The first quantity in the final objective \eqref{eq:3:lower_bound} resembles look-ahead planning. 
Thus, we approximate the optimal likelihood utilizing a shooting method. 
For the optimization of a parameterized distribution $q(\vbeta; \vtheta)$ we can either apply gradient-based~\cite{peters2010relative} or apply sampling-based techniques~\cite{urain_cep_2021, williams_mppi_2017, pinneri_icem_2021,williams_mppi_2017}. 
Since online sampling-based techniques are promising in practice, we chose the \acrshort{icem} approach \cite{ kobilarov_cemmpc_2012, pinneri_icem_2021}, i.e., after shooting, we take the $k$ best samples and apply moment matching to them. 
As we can calculate the entropy $\E[q(\vbeta)]{\log q(\vbeta)}$ in \eqref{eq:3:lower_bound} of our parameterized distribution $q(\vbeta; \vtheta)$ in closed form, it is a constant value applying \acrshort{icem}. 
Hence, we end up with the final objective for the iCEM method results in 
\begin{align*}
    \gJ_\text{CEM}(\vbeta) = 
    \mathbb{E}_{q(\vbeta)\hat{q}(\tau)}\big[\,
    &\textstyle\sum_{i=t}^{h}\log p(\gO_{t}{=}1\mid, \action_{t}, \state_{t})\\
    &+\textstyle\lambda_{\pi}\log \pi(\va_t\mid \vbeta, \vs_{t})~\big],
\end{align*} 
with regularization parameter $\lambda_{\pi}$.
After selecting the $k$ best samples, we estimate the mean and precision of the Dirichlet distribution separately \cite{minka_2000_dirichlet}. 
For the precision, a Newton-Raphson like method is applied, while for the mean, a fixed-point iteration takes place.
\section{EXPERIMENTS}
\begin{figure}[t!]
    \centering
    \includegraphics[width=\linewidth]{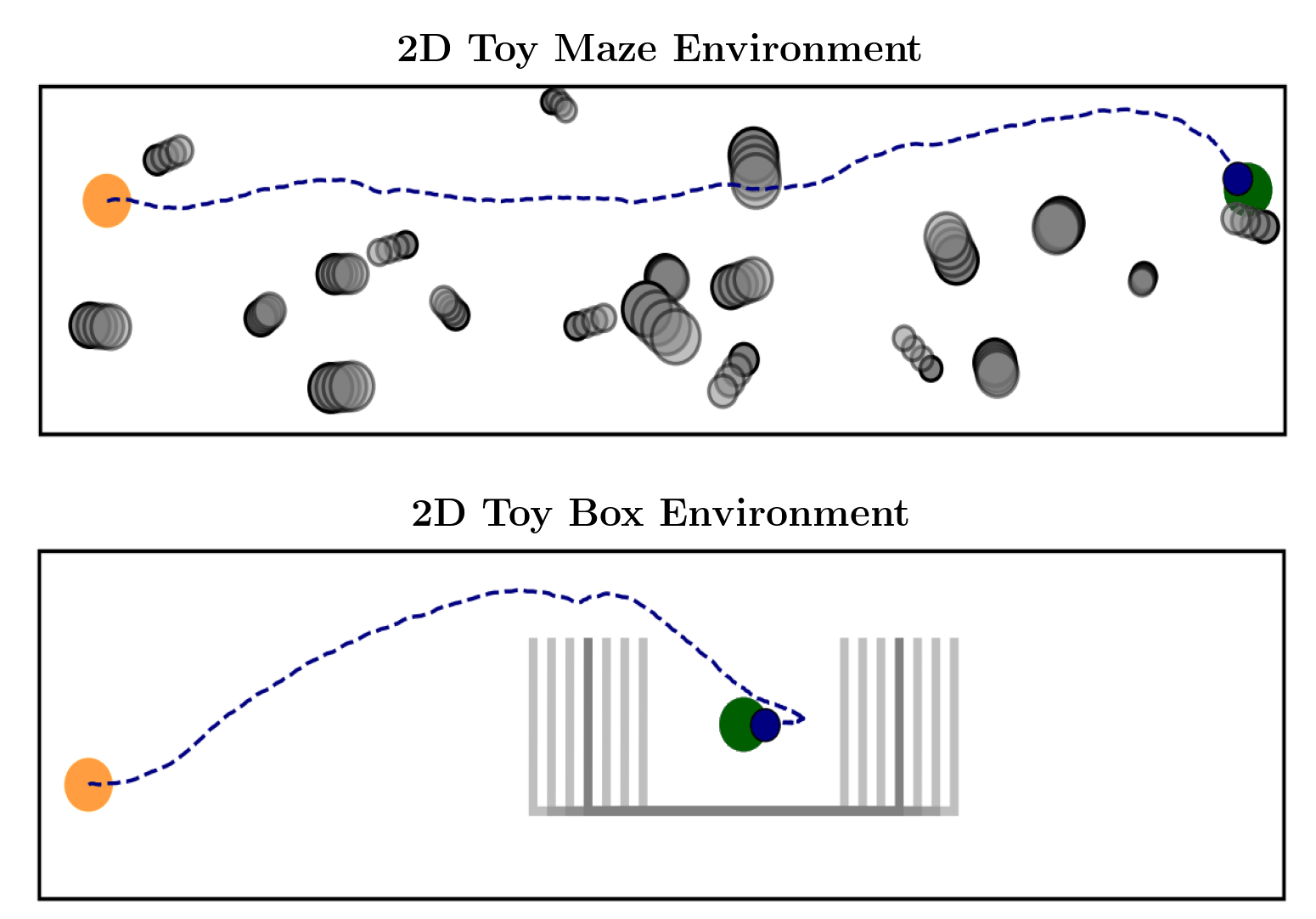}
    \caption{
    2D Toy environments for planar point-mass navigation. The orange dot denotes the start and the green one the goal location. \textbf{Top.} The toy maze environment with dynamic obstacles. \textbf{Bottom.} the toy box environment, in which a box moves horizontally at a constant speed. The goal is fixed in the center of the moving box.}
    \vspace{-0.5cm}
    \label{fig:4:box:toy:environments}
\end{figure}
In this section, we benchmark \acrshort{hipbi} against two baselines. The algorithm \acrfull{rmpflow}\,\cite{cheng_rmpflow_2018} works as a baseline from the family of reactive policies, that corresponds to a graph-based syntethis framework, and combines individual local \gls{rmp}s to generate global dynamical behavior.
A framework for real-time planning and second baseline is the \acrfull{mpcicem}\,\cite{pinneri_icem_2021} framework. The method utilizes \acrfull{icem} for trajectory optimization in a \acrfull{mpc} scheme. 
First, two low-dimensional planar navigation environments give us insights into how the different algorithms behave under different environmental conditions. Then, we consider a high-dimensional robotic simulation. This environment shows how \acrshort{hipbi} adapts to high-dimensional state and action spaces. In all experiments, we set $\lambda_\pi=0$.

\begin{figure*}[t]
\centering
\includegraphics[width=\linewidth, trim={0.75cm 0 1cm 0},clip]{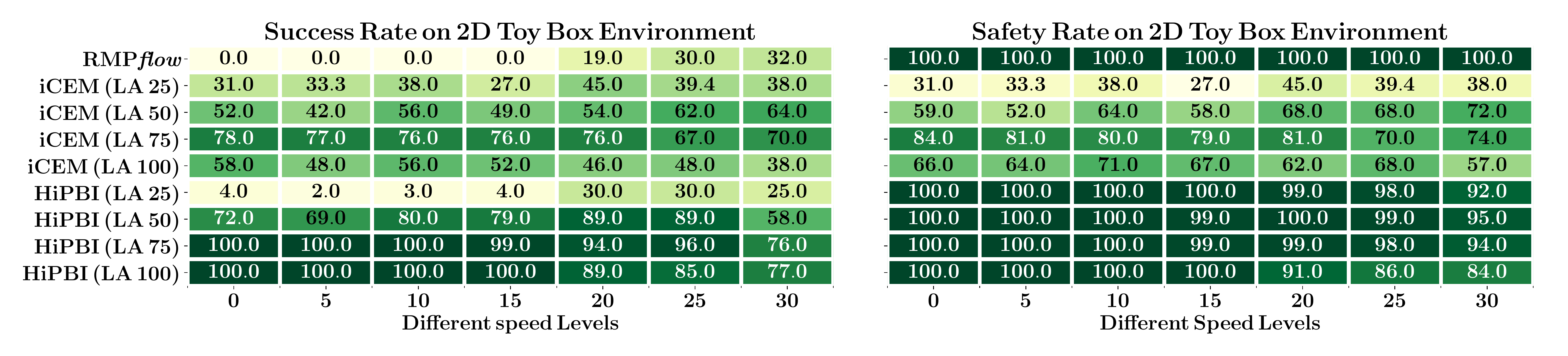}
\vspace{-0.75cm}
\captionof{figure}{
The results of an ablation study in the 2D toy box environment. The speed increases from a minimum of zero (static) to 30 pixels per step (dynamic). We compare the baselines, i.e., \acrshort{rmpflow} and \acrshort{mpcicem}, with our method \acrshort{hipbi} and employ different look-ahead horizons (LA). \textbf{Left.} The success rate shows the performance. \textbf{Right.} The safety rate indicates how often no collision occurs.}
\label{fig:4:box:toy:velocity}
\vspace{-0.5cm}
\end{figure*}
\begin{table*}[bp]
\vspace{-0.25cm}
\begin{center}
\captionof{table}{
Evaluation of the baselines, i.e., \acrshort{rmpflow} and \acrshort{mpcicem}, and our method \acrshort{hipbi} on the planar point-mass navigation tasks using different metrics: (i) the success rate (SUC); (ii) the safety rate (SAFE); (iii) the L2 distance from the final state to the goal (L2D); and (iv) the time steps required to reach the goal (TS). We employed different look-ahead horizons depicted as LA. The quantities S and A indicate whether the dynamics run synchronously with the algorithm or asynchronously. \textbf{Left.} Experiments took place in the 2D toy box environment. \textbf{Right.} Results in the 2D toy maze environment are highlighted.
}
\label{sec:4:tab:toy}
\resizebox{\textwidth}{!}{
\begin{tabular*}{\linewidth}{l
@{\extracolsep{\fill}}
c c c c @{\extracolsep{\fill}} c
c c c c}
\toprule
\phantom{Var.} &  
\multicolumn{4}{c}{2D Toy Box Environment} && \multicolumn{4}{c}{2D Toy Maze Environment}\\
\cmidrule{2-5}
\cmidrule{7-10} 
& {SUC$[\%]$} & {SAFE$[\%]$} & {L2D} & {TS} & {} & {SUC$[\%]$} & {SAFE$[\%]$} & {L2D} & {TS} \\
\midrule
RMP\textit{flow} 
& $0$ & $\mathbf{100}$ & $198.9\,\pm1.5$ & $500.0\,\pm0.0$ 
&
& $77$ & $89$ & $161.5\,\pm620.0$ & $330.7\,\pm191.3$ \\[1ex]
iCEM (LA 25, S) 
& $38$ & $\mathbf{100}$ & $77.2\,\pm91.2$ & $353.9\,\pm277.9$
&
& $98$ & $\mathbf{99}$ & $\mathbf{17.4\,\pm175.1}$ & $\mathbf{133.1\,\pm65.4}$ \\
iCEM (LA 50, S) 
& $57$ & $\mathbf{100}$ & $56.6\,\pm73.6$ & $271.0\,\pm282.5$ 
&
& $\mathbf{99}$ & $\mathbf{99}$ & $21.1\,\pm163.1$ & $167.0\,\pm59.8$ \\
iCEM (LA 75, S) 
& $90$ & $\mathbf{100}$ & $43.3\,\pm82.7$ & $195.2\,\pm177.2$
&
& $\mathbf{99}$ & $\mathbf{99}$ & $37.7\,\pm166.7$ & $223.8\,\pm90.5$ \\
\textbf{HiPBI} (LA 25, S) 
& $2$ & $\mathbf{100}$ & $189.3\,\pm44.7$ & $490.9\,\pm81.8$ 
&
& $98$ & $\mathbf{99}$ & $20.3\,\pm172.7$ & $247.6\,\pm55.8$ \\
\textbf{HiPBI} (LA 50, S) 
& $61$ & $\mathbf{100}$ & $49.5\,\pm75.6$ & $276.6\,\pm251.2$
&
& $\mathbf{99}$ & $\mathbf{99}$ & $17.5\,\pm162.6$ & $247.5\,\pm47.6$ \\
\textbf{HiPBI} (LA 75, S) 
& $\mathbf{100}$ & $\mathbf{100}$ & $7.3\,\pm5.9$ & $\mathbf{131.9\,\pm18.0}$ 
&
& $\mathbf{99}$ & $\mathbf{99}$ & $19.0\,\pm171.7$ & $252.1\,\pm47.3$ \\[1ex]
iCEM (LA 25, A) 
& $31$ & $31$ & $95.5\,\pm94.8$ & $372.9\,\pm265.6$
&
& $40$ & $40$ & $409.6\,\pm570.0$ & $356.6\,\pm245.5$ \\
iCEM (LA 50, A) 
& $54$ & $64$ & $64.3\,\pm83.2$ & $292.5\,\pm271.7$ 
&
& $4$ & $4$ & $774.0\,\pm436.1$ & $488.9\,\pm78.1$ \\
iCEM (LA 75, A) 
& $79$ & $85$ & $63.1\,\pm84.4$ & $237.7\,\pm208.6$ 
&
& $0$ & $0$ & $974.1\,\pm287.5$ & $499.2\,\pm22.4$ \\
\textbf{HiPBI} (LA 25, A) 
& $7$ & $\mathbf{100}$ & $178.6\,\pm71.1$ & $477.1\,\pm120.3$ 
&
& $83$ & $84$ & $116.2\,\pm386.3$ & $294.2\,\pm131.4$ \\
\textbf{HiPBI} (LA 50, A) 
& $73$ & $\mathbf{100}$ & $40.1\,\pm76.9$ & $324.3\,\pm169.7$ 
&
& $85$ & $\mathbf{87}$ & $\mathbf{100.0\,\pm357.9}$ & $\mathbf{293.4\,\pm123.7}$ \\
\textbf{HiPBI} (LA 75, A) 
& $\mathbf{100}$ & $\mathbf{100}$ & $\mathbf{8.5\,\pm6.0}$ & $\mathbf{205.8\,\pm35.3}$
&
& $\mathbf{86}$ & $\mathbf{87}$ & $106.1\,\pm376.5$ & $297.3\,\pm122.1$ \\
\bottomrule
\end{tabular*}}%
\end{center}
\end{table*}
\paragraph*{\textbf{Toy Environments}} 
The \gls{2dmaze} is a dynamic 2D planar environment on which a particle navigates from a random start to a random goal position (see Fig.~\ref{fig:4:box:toy:environments}). In the environment, a given number of $\nobstacles$ circular obstacles -- partly static, partly dynamic -- bar the way. The environment randomly sets obstacles inside a restricted area between the start (orange point) and goal positions (green point). We model the movement of the obstacles using a constant velocity model. 
\gls{2dmaze} mimics a dense, cluttered and dynamic environment. 

Unlike \gls{2dmaze}, the \gls{2dbox} presents a dynamic domain in which constant local optima exist (see Fig.~\ref{fig:4:box:toy:environments}). The box is dynamic, and its motion is modeled as a constant velocity one. The start position is sampled randomly to the right or left of the box. The challenge comes from local optima under the box or in front, i.e., to the left or the right. Furthermore, the dynamic nature complicates the planning of a feasible solution. 
Validation is important as local optima are constantly changing in \gls{2dmaze} -- they appear and disappear independently - whereas they exist permanently in \gls{2dbox}. Therefore, \gls{2dbox} shows us the effectiveness in overcoming constant local optima.

We consider four different metrics for validation in \gls{2dmaze} and \gls{2dbox}: (i) the success rate (SUC), indicating the percentage of times the goal has been reached; (ii) the safety rate (SAFE), implying collision-free motions; (iii) the final l2 distance (L2D) to the goal; and (iv) the needed time steps (TS) until the goal is reached. As \acrshort{rmpflow} runs at high frequency, we studied \acrshort{mpcicem} and \acrshort{hipbi} in a synchronous (S) and asynchronous (A) mode. 
In the former, \acrshort{mpcicem} and \acrshort{hipbi} have sufficient time to find feasible solutions -- as the environment remains fixed during the planning. 
In the latter, planning runs asynchronously with the environment, and online planning is needed. 
Therefore, we applied different look ahead (LA) horizons for both the basic \acrshort{mpcicem} and the \acrshort{hipbi} methods. 

\acrshort{rmpflow} employs the composition of experts with attractive and repulsive forces. To achieve a curling behavior, we choose an expert ${\pi_{\textrm{curl}}\,\bot\,\pi_{\textrm{goal}}}$ that exerts forces normal to the goal attractive force. $\pi_{\textrm{curl}}$ scales proportional to $\pi_{\textrm{goal}}$ and, thus, vanishes if the particle reaches the goal. We apply two mutually balancing agents to avoid constant rotational forces, i.e., ${\pi_{\textrm{curl}_{i}}= -\pi_{\textrm{curl}_{j}}}$. Although this extension does not affect \acrshort{rmpflow}, \acrshort{hipbi} adapts the weights and hence achieves curling behavior.

\begin{figure}[!b]
    \centering
    \vspace{-0.55cm}
    \includegraphics[width=\linewidth, height=4.5cm, trim={1.25cm 0 1cm 0},clip]{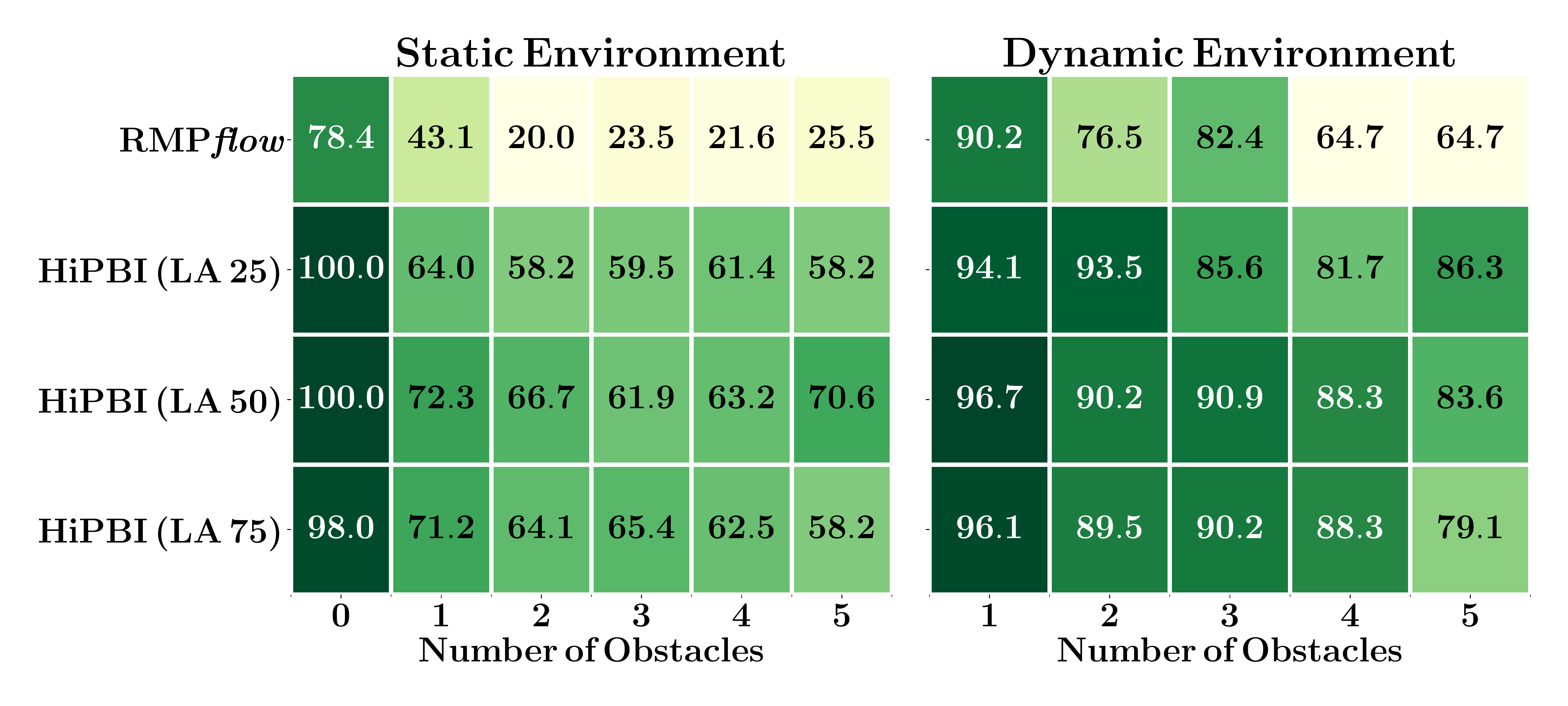}
    \vspace{-0.7cm}
    \caption{Evaluation study on the manipulation environment. We benchmark our approach \gls{hipbi} in a static and a dynamic setting against the baseline \gls{rmpflow}. \textbf{Left.} The success rate in a static environment. The number of obstacles varies from zero up to a maximum of five. \textbf{Right.} The success rate in a dynamic environment. A maximum of five movable obstacles are used.}
    \vspace{-0.5cm}
    \label{fig:4:eval:panda}
\end{figure}

Table~\ref{sec:4:tab:toy} summarizes our results. 
In \gls{2dbox}, \acrshort{rmpflow} converges to local optima. 
In synchronous mode, \acrshort{mpcicem} and \acrshort{hipbi} improve their success rate with increasing look-ahead size and encounter no collisions. 
However, \acrshort{mpcicem} loses performance in asynchronous mode as it cannot react fast to environmental changes. 
\acrshort{hipbi}, which combines online planning with reactive control, achieves a 100\% success and safety rate with a look-ahead of 75 without suffering from the same performance gap.
In \gls{2dmaze}, \acrshort{rmpflow} achieves a success rate of 77\%. 
As in \gls{2dbox}, \acrshort{mpcicem} and \acrshort{hipbi} perform well in synchronous mode. 
However, in the asynchronous mode, we notice that \acrshort{mpcicem} suffers again from its slow response to environmental changes, more noticeable at higher look-ahead values. Interestingly, this behavior is reversed in \gls{2dbox}. 
Dodging under or over one box is simpler than avoiding multiple particles. 
The performance of \acrshort{hipbi} improves with increasing look-ahead planning. 
The similar collision rate as \acrshort{rmpflow} is reasonable as both use the same parameters for the underlying \gls{rmp}.
Due to planning, \acrshort{hipbi} improves the success rate compared to the myopic \acrshort{rmpflow}.
\vspace{-0.05em}
In Figure~\ref{fig:4:box:toy:velocity}, we show a comparison on \gls{2dbox} with different speed levels of the box. 
This comparison provides insights into the responsiveness to environmental changes. 
We see how success and safety rates change as velocity increases. 
Regardless of the velocity, \acrshort{rmpflow} does not collide with the box environment but has a low success rate. 
\acrshort{mpcicem} achieves a higher success rate, but collisions occur more often. 
The ablation study highlights how \acrshort{hipbi} combines the advantages of low-level reactiveness and high-level planning. 
Increasing speed has a small influence on \acrshort{hipbi}, with only a slight drop in performance at a speed level of 30 pixels per step while maintaining a sufficient safety rate.
\begin{figure*}[h!]
    \centering
    \includegraphics[width=\textwidth]{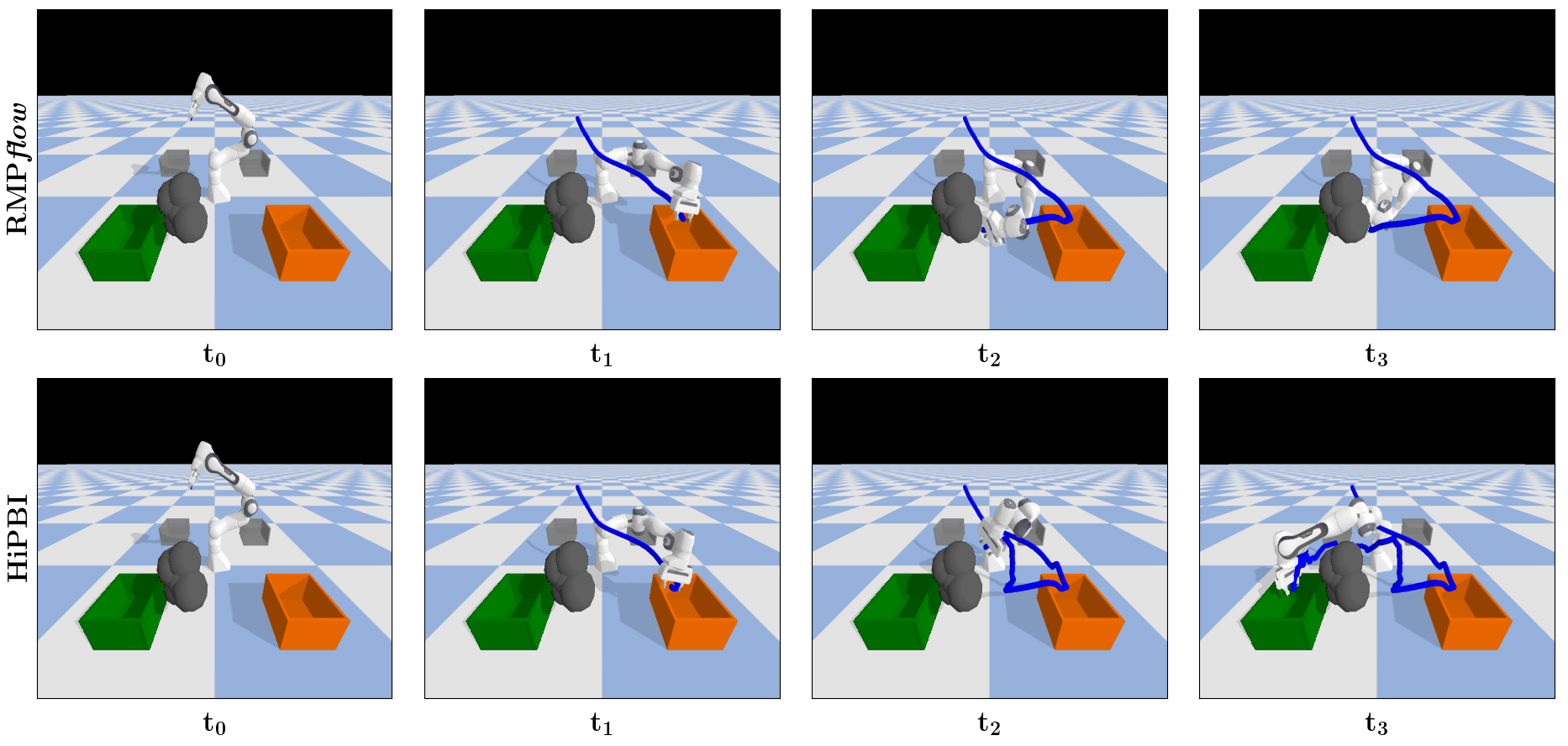}
    \vspace{-0.65cm}
    \caption{Manipulation environment in which the intermediate (orange) and target (green) boxes are randomly selected out of four boxes. Five randomly generated grey obstacles obstruct the path of the 7DoF manipulator robot. With blue, we denote the executed trajectory. \textbf{Top.} Performance of \gls{rmpflow} method that gets stuck in a local optimum. \textbf{Bottom.} Performance of our proposed \gls{hipbi}, that successfully discovers an obstacle-free path to the target.}
    \label{fig:4:sequence:panda}
    \vspace{-0.65cm}
\end{figure*}
\paragraph*{\textbf{Manipulation Environment}} 
We investigate the performance of \acrshort{hipbi} on a high-dimensional robotics task with a 7DoF manipulator in the physics engine PyBullet~\cite{benelot_2018_pybullet}. Fig.~\ref{fig:4:eval:panda} shows the arm surrounded by four boxes. In each round, the robot must first get to a randomly selected intermediate goal (orange box). After it reaches the intermediate goal, it has to reach the final goal (green box). Several sphere-like objects block the way during the path from the intermediate to the goal box. Thus, this task resembles a high-dimensional pick-and-place task involving multiple local optima. 

\vspace{-0.05em}
In an ablation study, Fig.~\ref{fig:4:sequence:panda}, we compared the performance of \acrshort{hipbi} against \acrshort{rmpflow}. Long horizon planning was not feasible with \acrshort{mpcicem}, thus we do not add any comparison, as the short horizons did not give satisfactory results.
We consider two modes, a static one and a dynamic one. 
In the former, zero to five static spheres are randomly sampled in a predefined space between the intermediate and goal box. 
In the latter, we use one to five dynamic spheres similar to the former case assuming constant velocity models.
The dynamic obstacles are restricted to staying within the path of the panda and goal box.
\acrshort{rmpflow} uses eight experts, which included self-collision avoidance, joint and velocity limitations, goal-reaching, and avoidance of obstacles such as floors, boxes, and spheres.
Unlike in \gls{2dbox} and \gls{2dmaze}, \acrshort{hipbi} leverages the same experts and omits the local curling policies.
Our approach adjusts the importance of the experts to achieve desired dynamical behavior. 
Due to the computational complexity of planning algorithms and the advantage of optimizing in parameter space, we apply the \acrshort{hipbi} in asynchronous mode -- as 2D results confirm our assumption. 
While the high-level planner optimizes at a lower frequency, the local policies ensure reactive behavior. 
By choosing a Dirichlet distribution, we guarantee that each expert affects the dynamic system. Thus, no scenarios arise in that local policies are switched off.

In Figure~\ref{fig:4:eval:panda}, we present an ablation study and see that \acrshort{rmpflow} performs better in a dynamic setting, strengthening our assumption that local optima alter in dynamic environments. 
However, the success rate decreases with an increasing number of obstacles. 
This outcome is reasonable, as each obstacle induces another constraint creating more local optima. 
\acrshort{hipbi} demonstrates significantly improved results with a look-ahead horizon of $25$, corresponding to $2.5\,s$ at a planning frequency of $10\,Hz$. In the static environment with fixed local optima, \acrshort{hipbi} also outperforms \acrshort{rmpflow}.

Figure~\ref{fig:4:sequence:panda} compares two executed trajectories of \acrshort{rmpflow} (top) and \acrshort{hipbi} (bottom) at four different time points. 
\acrshort{rmpflow} naturally follows the myopic behavior and ends in a local optimum. \acrshort{hipbi}, on the other hand, exploits the information of the hierarchical high-level planning scheme.
The reactive leaping motion indicates the low-level \gls{rmp}s. High-level planning enables feasible solutions avoiding local optima and reaching the goal. 


\section{CONCLUSIONS}
We presented \acrfull{hipbi}, a method for reactive motion generation that combines, at the low level, myopic reactive motion policies that can be modeled as a \acrfull{poe}, and, on the high level, a sampling-based online planner on the parameter space of the policies, that decides over the optimal weighting of the experts. Our method dynamically adapts the importance of the different policies and shows superior performance in terms of task success-rate and safety (in terms of collision avoidance) against representative baselines, as demonstrated both in complex planar environments, and in high-dimensional robotic manipulation tasks in face of clutter and dynamic changes.

As our method comes with the cost of higher computational complexity, we will explore collocation methods for the planning process in the future and apply the approach to appropriately designed real robot environments. Furthermore, the probabilistic inference framework assumes a prior distribution to provide prior knowledge to the system. 
We will discuss the form and realization of such a prior, e.g., by imitation or offline reinforcement learning.

\clearpage
\bibliographystyle{source/IEEEtran}
\bibliography{source/IEEEabrv,mybib_tidy}
\end{document}